\documentclass[conference]{IEEEtran}
\IEEEoverridecommandlockouts
\usepackage{cite}
\usepackage{amsmath,amssymb,amsfonts}
\usepackage{algorithmic}
\usepackage{graphicx}
\usepackage{textcomp}
\usepackage{xcolor}
\usepackage{physics}
\usepackage{url}
\usepackage[font=footnotesize,skip=1pt]{caption}  

\def\BibTeX{{\rm B\kern-.05em{\sc i\kern-.025em b}\kern-.08em
    T\kern-.1667em\lower.7ex\hbox{E}\kern-.125emX}}
\begin{document}

\title{Quantum-Enhanced Classification of Brain Tumors 
Using DNA Microarray Gene Expression Profiles}

\author{\IEEEauthorblockN{Emine Akpinar}
\IEEEauthorblockA{\textit{Department of Physics} \\
\textit{Yildiz Technical University}\\
İstanbul, Türkiye \\
emineakpinar28@gmail.com}
\and
\IEEEauthorblockN{Batuhan Hangun}
\IEEEauthorblockA{\textit{Department of Computer Engineering} \\
\textit{Yildiz Technical University}\\
İstanbul, Türkiye \\
batuhanhangun@gmail.com}
\and
\IEEEauthorblockN{Murat Oduncuoglu}
\IEEEauthorblockA{\textit{Department of Physics} \\
\textit{Yildiz Technical University}\\
İstanbul, Türkiye \\
murato@yildiz.edu.tr}
\and
\IEEEauthorblockN{Oguz Altun}
\IEEEauthorblockA{\textit{Department of Computer Engineering} \\
\textit{Yildiz Technical University}\\
İstanbul, Türkiye \\
oaltun@yildiz.edu.tr}
\and
\IEEEauthorblockN{Onder Eyecioglu}
\IEEEauthorblockA{\textit{Computer Engineering Department} \\
\textit{Bolu Abant Izzet Baysal University}\\
Bolu, Türkiye \\
onder.eyecioglu@ibu.edu.tr}
\and
\IEEEauthorblockN{Zeynel Yalcin}
\IEEEauthorblockA{\textit{Department of Physics} \\
\textit{Yildiz Technical University}\\
İstanbul, Türkiye \\
zyalcin@yildiz.edu.tr }
}

\maketitle

\begin{abstract}
DNA microarray technology enables the simultaneous measurement of expression levels of thousands of genes, thereby facilitating the understanding of the molecular mechanisms underlying complex diseases such as brain tumors and the identification of diagnostic genetic signatures. To derive meaningful biological insights from the high-dimensional and complex gene features obtained through this technology and to analyze gene properties in detail, classical artificial intelligence (AI)-based approaches such as machine learning (ML) and deep learning (DL) are widely employed. However, these methods face various limitations in managing high-dimensional vector spaces and modeling the intricate relationships among genes. In particular, challenges such as hyperparameter tuning, computational costs, and high processing power requirements can hinder their efficiency. To overcome these limitations, quantum computing and quantum AI approaches are gaining increasing attention. Leveraging quantum properties such as superposition and entanglement, quantum methods enable more efficient parallel processing of high-dimensional data and offer faster and more effective solutions to problems that are computationally demanding for classical methods. In this study, a novel model called “Deep VQC” is proposed, based on the Variational Quantum Classifier (VQC) approach. Developed using microarray data containing 54,676 gene features, the model successfully classified four different types of brain tumors—ependymoma, glioblastoma, medulloblastoma, and pilocytic astrocytoma—alongside healthy samples with high accuracy. Furthermore, compared to classical ML algorithms, the Deep VQC model demonstrated either superior or comparable classification performance. These results highlight the potential of quantum AI methods as an effective and promising approach for the analysis and classification of complex structures such as brain tumors based on gene expression features.
\end{abstract}


\section{Introduction}
Brain tumors, or intracranial neoplasms, arise as a result of the abnormal growth of cells within or around the brain, and they constitute one of the leading causes of increasing mortality rates across different age groups. When this tumor formation originates from the brain tissue itself, it is classified as primary; if it originates from another part of the body and represents a distant metastasis of a primary tumor, it is considered secondary. Accurate diagnosis and evaluation of these tumors are critically important for both treatment planning and prognosis. The diagnosis and grading of brain tumors largely rely on radiological findings and data derived from the histopathological characteristics of the tumors. In addition, technologies such as DNA microarrays enable the identification of gene expression profiles specific to brain tumors, thus allowing molecular-level classification and the development of more effective and precise treatment strategies \cite{shu2018}.

DNA microarray (or simply microarray) is a molecular biology technique used to determine the overall gene expression profiles of various biological entities—such as organisms, tissues, or cells—and to perform mutation analysis. In other words, it is a method that simultaneously monitors the expression levels of thousands of genes in order to identify which genes are expressed or activated, and to what extent, under specific conditions \cite{cho2003machine, blohm2001,
kagawa2006}. This technique has accelerated the transition from gene-level research to genome-wide studies, making it possible to assess expression levels across the entire genome. Therefore, rather than analyzing individual genes to reveal gene expression patterns involved in complex biological processes such as tumor development, the ability to examine many—or even all—genes at once is crucial for understanding disease mechanisms and for developing more effective diagnostic and therapeutic strategies.

While microarray technology offers significant advantages, it also involves a complex analytical process that presents several challenges. One of the primary challenges is the enormous volume of data generated by microarray experiments, often referred to as the high-dimensionality problem. For instance, a microarray experiment involving 15,000 genes related to a brain tumor yields data that can be represented as a 15,000-dimensional vector in a classical vector space. This high dimensionality poses substantial computational difficulties when analyzing such datasets.

Another major challenge lies in extracting meaningful biological insights from this vast and complex microarray data \cite{sharm2012}. Identifying gene features that may be critical for brain tumor classification and early diagnosis presents significant obstacles, both in terms of computational cost and data analysis.
In addition to these challenges, microarray datasets typically contain a relatively small number of samples (or cases) while encompassing thousands of gene features. As a result, a vast amount of gene expression information is obtained from each individual case. This imbalance is known in data science as the "curse of dimensionality" and represents one of the most significant challenges when working with microarray data \cite{gupta2022, liu2011}.
To address these challenges and effectively analyze the large-scale data generated by microarray experiments, classical artificial intelligence (AI) approaches—particularly machine learning (ML) and deep learning (DL)—have been increasingly employed in recent years. Through these methods, microarray data can be analyzed to extract and interpret meaningful features or patterns. In ML-based approaches, gene expression profiles—represented as high-dimensional vectors—can be processed using feature selection techniques to identify the most relevant and informative gene features, thereby reducing both data dimensionality and computational cost. Moreover, ML algorithms can be trained on the selected features to classify new samples through supervised learning. In DL approaches, which rely on hierarchical feature learning, features are automatically extracted from raw data, enabling the learning of high-level representations directly from microarray inputs. In the literature, both ML and DL methods have been successfully applied to various tasks involving microarray data, including tumor and cancer classification \cite{torkey2021, joshi2024}, biomarker discovery \cite{chang2020}, and understanding disease mechanisms \cite{chang2020}.

However, despite their promising performance in selecting key gene features and classifying complex diseases such as brain tumors, these methods face certain limitations. Specifically, they are affected by high-dimensional data structures involving thousands of gene features, complex relationships between features, the dimensionality of the classical vector space, and the need for careful hyperparameter tuning. Furthermore, as data volume increases, these methods tend to require significantly higher computational resources.

Recent studies have shown that in order to overcome data- and computation-based challenges and to address complex problems such as brain tumor classification based on gene features, there is a growing need for quantum computing and quantum computing-based AI techniques.

Quantum computing refers to the development of computational technologies grounded in the fundamental principles of quantum mechanics, aiming to address problems that exceed the capabilities of classical computational methods. This approach exploits quantum phenomena such as superposition, entanglement, and interference to achieve solutions where classical methods either fall short or become computationally infeasible \cite{gill2022}. In some cases, quantum algorithms can offer results comparable to classical ones, while in others they promise significantly higher processing capacity and computational speed. Moreover, this paradigm enables the processing, analysis, and transmission of information within a quantum framework.

Quantum computing-based AI, on the other hand, seeks to harness the power of quantum computation in combination with classical AI algorithms to enhance performance in tasks such as pattern recognition, data processing, classification, and optimization. Currently, quantum computing and  quantum computing-based AI methods are being explored in various applications, including microarray data analysis, gene selection, and the early diagnosis of brain tumors \cite{saggi2024multiomic, akpinar2024, wang2025, akpinar2022}.

In this study, a classification model called 'Deep Variational Quantum Classifier' (Deep VQC) is proposed, based on a quantum AI model, the Variational Quantum Classifier (VQC). The model successfully differentiates four distinct types of brain tumors (ependymoma, glioblastoma, medulloblastoma, and pilocytic astrocytoma) as well as normal samples from healthy individuals, using microarray data. Despite the limitations of noisy intermediate-scale quantum (NISQ) computers, the proposed model works in harmony with NISQ devices, achieving high classification accuracy with fewer parameters and lower circuit depth.

\section{Materials and Methods}
The proposed model for classifying four different brain tumor types and normal samples based on microarray data consists of two stages: a classical part and a quantum part. The first stage involves dataset definition and preprocessing steps, while the quantum part outlines the steps of the proposed Deep VQC model. 

\subsection{Classical Part}
\subsubsection{Dataset and Preprocessing}
\begin{figure}
    \centering
   \includegraphics[width=0.7\columnwidth,trim={3pt 2pt 3pt 5pt},clip]{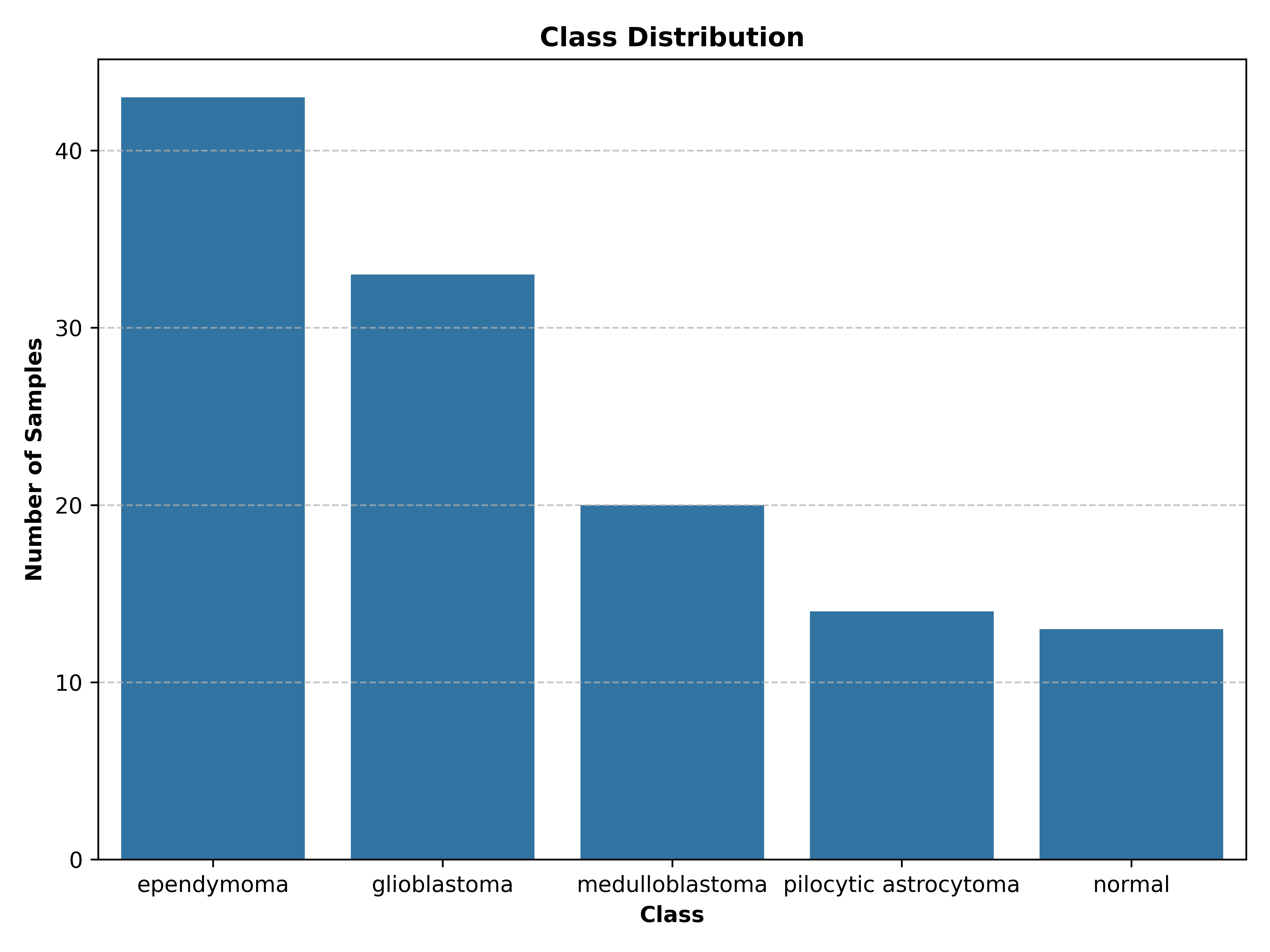}
    \caption{Class distribution of the microarray dataset, which includes samples from four different brain tumor types (ependymoma, glioblastoma, medulloblastoma, and pilocytic astrocytoma), as well as a normal class representing healthy individuals.}
    \label{fig:class_dist}
\end{figure}

The microarray dataset used in this study is a publicly available dataset obtained from the Curated Microarray Database (CuMiDa) platform \cite{feltes2019}. CuMiDa is a large-scale data repository containing 78 cancer microarray datasets of Homo sapiens, derived from the examination of over 30,000 microarray experiments available in the Gene Expression Omnibus (GEO). The microarray dataset obtained from this platform includes four different types of brain tumors (ependymoma, glioblastoma, medulloblastoma, and pilocytic astrocytoma) as well as a normal class for healthy individuals. Additionally, the dataset consists of samples from 130 individuals, with each sample containing 54,676 gene features. A graphical representation of the class distribution is shown in Fig. \ref{fig:class_dist}.

\subsubsection{Normalization and Dimension Reduction (with and without)}
Before proceeding to the classification step with the Deep VQC model, one of the most important operations to be applied on the data is data normalization, which is performed to bring the data within a specific range. This step is particularly crucial in quantum AI studies to maintain the consistency of quantum states. In this study, the data was first balanced in terms of mean (subtracting the mean from each value) and standard deviation (dividing the obtained difference by the standard deviation) using the Standard Scaler. Then, the Min-Max Scaler was applied to scale all values to the range of [0, 1].
The next step consists of two phases. First, the normalized microarray data was directly fed into the proposed Deep VQC model, enabling the classification of four brain tumor types and normal samples while preserving the original 54,676 gene features.
To make a comparison, Principal Component Analysis (PCA) was applied in the subsequent step to reduce the dimensions of the normalized data. During PCA, components were selected to retain 95\% of the total variance, thereby significantly reducing the data size while minimizing information cost.

\subsection{Quantum Part}
\begin{figure*}[t]
    \centering
    \includegraphics[width=\linewidth]{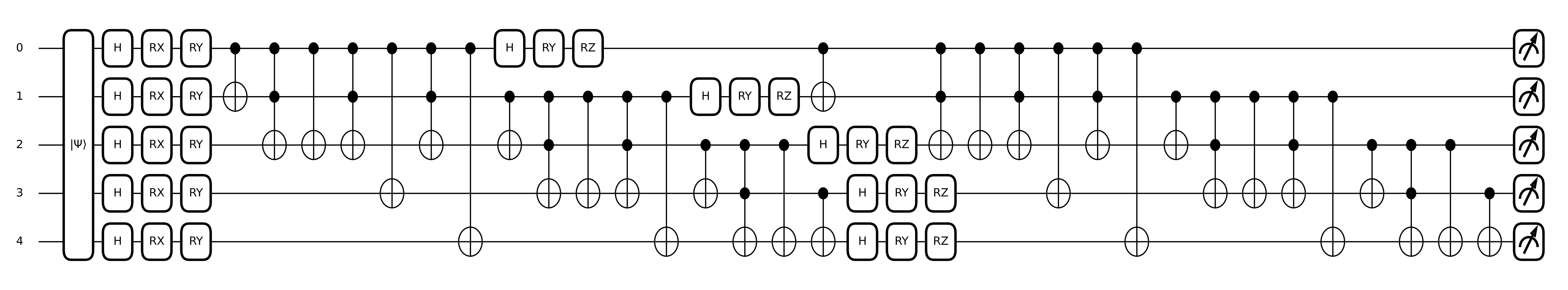}
    \caption{Representation of the proposed Deep VQC model. Due to space constraints, only the first layer is shown. The complete model consists of 25 layers.}
    \label{fig:qu_circ}
\end{figure*}

The Variational Quantum Classifier (VQC) model leverages a hybrid quantum-classical optimization loop to train the circuit parameters and determine their optimal values \cite{cerezo2021}. This method has become widely preferred in recent years for solving many problems and achieving quantum advantage on NISQ devices. In this context, the proposed Deep VQC model for brain tumor classification based on microarray data consists of feature mapping, two different Hardware Efficient Ansatz (HEA), measurement, and optimization steps. In this study, the quantum codes were executed on the default.qubit state-vector simulator provided by the PennyLane framework (version 0.39.0). The circuit structure of the Deep VQC model is shown in Fig. \ref{fig:qu_circ}.

\subsubsection{Feature Mapping}
In gate-based quantum computers, for classical data to be processed within a quantum circuit (such as for information transfer), it must be represented as a quantum state in Hilbert space. This transformation is achieved through the use of a feature map. Initially, all quantum states are typically expressed as product states in the form of \(\ket{0\dots0}\), allowing each qubit to be individually controlled and facilitating the state preparation process. In this study, the amplitude encoding technique was employed to represent normalized microarray data as quantum states. This technique is particularly advantageous for high-dimensional data, as it offers a qubit-efficient representation and thus enables more efficient data loading into quantum computers \cite{schuld2021machine}.

In the amplitude encoding method, classical input \(x=\ \left({x_0,\ x}_1,\ x_2,\ldots\ldots.,x_{2^n-1}\right)\) are expressed as the amplitudes of quantum states:
\begin{equation}
    \ket{\psi} = x_{0}\ket{00\dots0}+x_{1}\ket{00\dots1}+\cdots+x_{2^n-1}\ket{11\dots1}
\end{equation}
where \(x_{0}, x_{1}, x_{2}, \dots \; , x_{2^n-1}\) are normalized vectors.

\subsubsection{Hardware Eﬃcient Ansatz (HEA)}
Fundamentally, a hardware-efficient ansatz refers to a circuit architecture designed for general-purpose use rather than being tailored to a specific problem. It is characterized by a shallower depth and fewer circuit elements \cite{bharti2022}. Typically, such ansatz may lead to training challenges and prolonged convergence times when input parameters are randomly initialized. One of the key features of a Hardware-Efficient Ansatz (HEA) is its ability to incorporate circuit symmetries and optimize circuit depth by taking qubit correlations into account. Thus, an HEA is a circuit model that, after a feature mapping process, applies a set of parameterized (or non-parameterized) single-qubit and multi-qubit quantum gates to data expressed as quantum states, allowing to transform the input state into another quantum state on one or more qubits.

In general, a Hardware-Efficient Ansatz (HEA) is expressed as:
\begin{equation}
    \ket{\psi(\theta)}= U^{(L)}\theta^{(L)} \; \dots \; U^{(1)}\theta^{(1)}U^{(0)}\theta^{(0)}\ket{\psi_0}
\end{equation}
Here, \(\ket{\psi_0}\) represents the initial state, \(\theta=[\theta^{(0)},\theta^{(1)} \; \dots \; \theta^{(L-1)},\theta^{(L)}]\) are the variational or learnable parameters, and \(U^{k}\) represents the \textit{k}-th layer, which consists of a repetition of parameterized single-qubit gates and fixed multi-qubit entangling gates \cite{cao2019}.

In this study, two different HEAs have been proposed that operate sequentially, enabling the model to learn more significant and meaningful features from input microarray data expressed as quantum states. The first HEA consists of Hadamard gates, parameterized single-qubit RX and RY rotations applied to each qubit, followed by entangling operations using CNOT and Toffoli gates between qubit pairs. The second HEA, designed to enrich the learned representation, employs Hadamard gates, parameterized RY and RZ rotations, and the same pattern of entangling gates. This layered structure allows the model to capture complex quantum correlations within the input data.

\subsubsection{Measurement and Optimization}
After a series of unitary operations were performed on quantum states using two different HEAs, a measurement process was carried out on the qubits to obtain class probabilities and class predictions. At this stage, the measurements were performed in the Pauli-Z basis, and the expected values of five different qubits corresponding to five classes were calculated. Based on the obtained values, the probability distribution for each class was determined using the softmax function. This process can be expressed as follows:

\begin{equation}
    P_j = \frac{exp(\bra{\psi_j}\sigma_z\ket{\psi_j})}{\sum_{k=1}^{5}exp(\bra{\psi_k}\sigma_z\ket{\psi_k})}
\end{equation}

After the softmax function is applied, the class with the highest probability value among the five obtained probabilities is selected as the model's prediction. Additionally, the difference between the true class values \((y_j)\) and the predicted class values by the Deep VQC model \((\widehat{y_j})\) is computed using the cross-entropy cost function given by \eqref{eq:cross_ent}:

\begin{equation}
    C(\theta) = -\sum_{j=1}^{5}y_jlog((p_j)\theta),
\label{eq:cross_ent}
\end{equation}

To determine the optimal circuit parameters \(\theta\) that minimize the cost function, the Gradient Descent Optimizer method was used. Steps 2 and 3 are continuously repeated until the optimal values for the parameters in the HEAs are determined.

\section{Results}
In this study, a deep variational quantum classification model, referred to as Deep VQC, was proposed to distinguish between four different brain tumor types and normal samples based on microarray data. In the Deep VQC model, the number of qubits was fixed at 15, and the number of circuit layers was set to 25. In this way, the microarray data was effectively represented in the quantum environment, while increasing the number of layers in the HEA circuits enabled the learning of important features from the input data.

The results of the proposed model are presented under three different headings. Firstly, the 54,676 gene features from the original dataset were transferred to the quantum model, and brain tumor classification was performed. The results of this process are presented in the first section. In the next step, Principal Component Analysis (PCA) was applied to the microarray dataset during the preprocessing stage, and components were selected to preserve 95\% of the total variance. The results of this process are presented in the second section. In the third section, the results of classical machine learning (ML) algorithms, used to compare with the proposed Deep VQC model, are provided. The CuMiDa database includes 3-fold cross-validation results for various ML models used for the 5 different classes. These results are presented in this section, and the accuracy results of the Deep VQC model are compared with these results.

\subsection{Performance of the Deep VQC Model }
\begin{figure}[b]
    \centering
    \includegraphics[width=0.8\columnwidth]{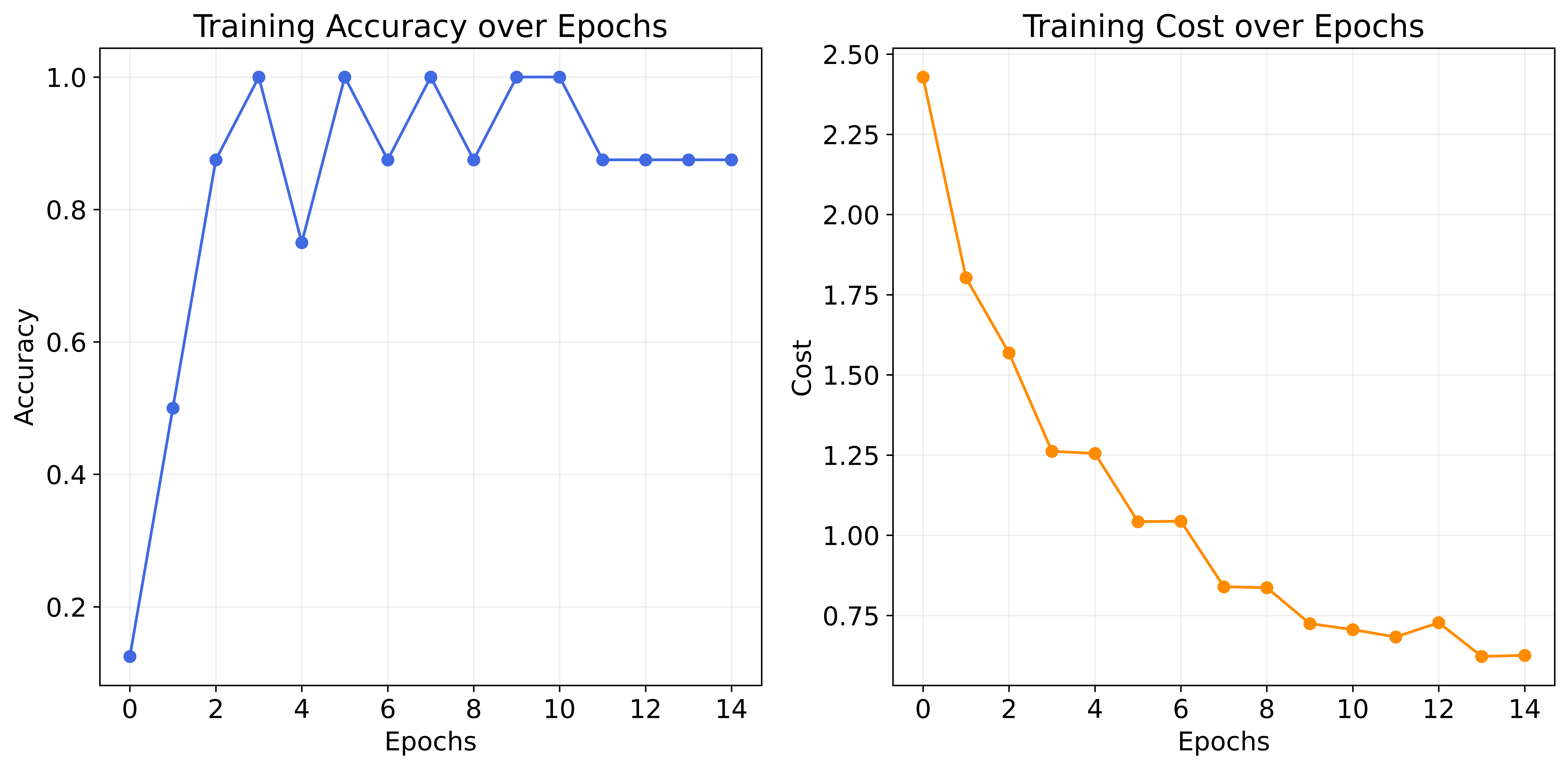}
    \caption{Training accuracy and cost curves of the Deep VQC model.}
    \label{fig:train_acc_1}
\end{figure}

\begin{figure}[t]
    \centering
    \includegraphics[width=0.8\columnwidth]{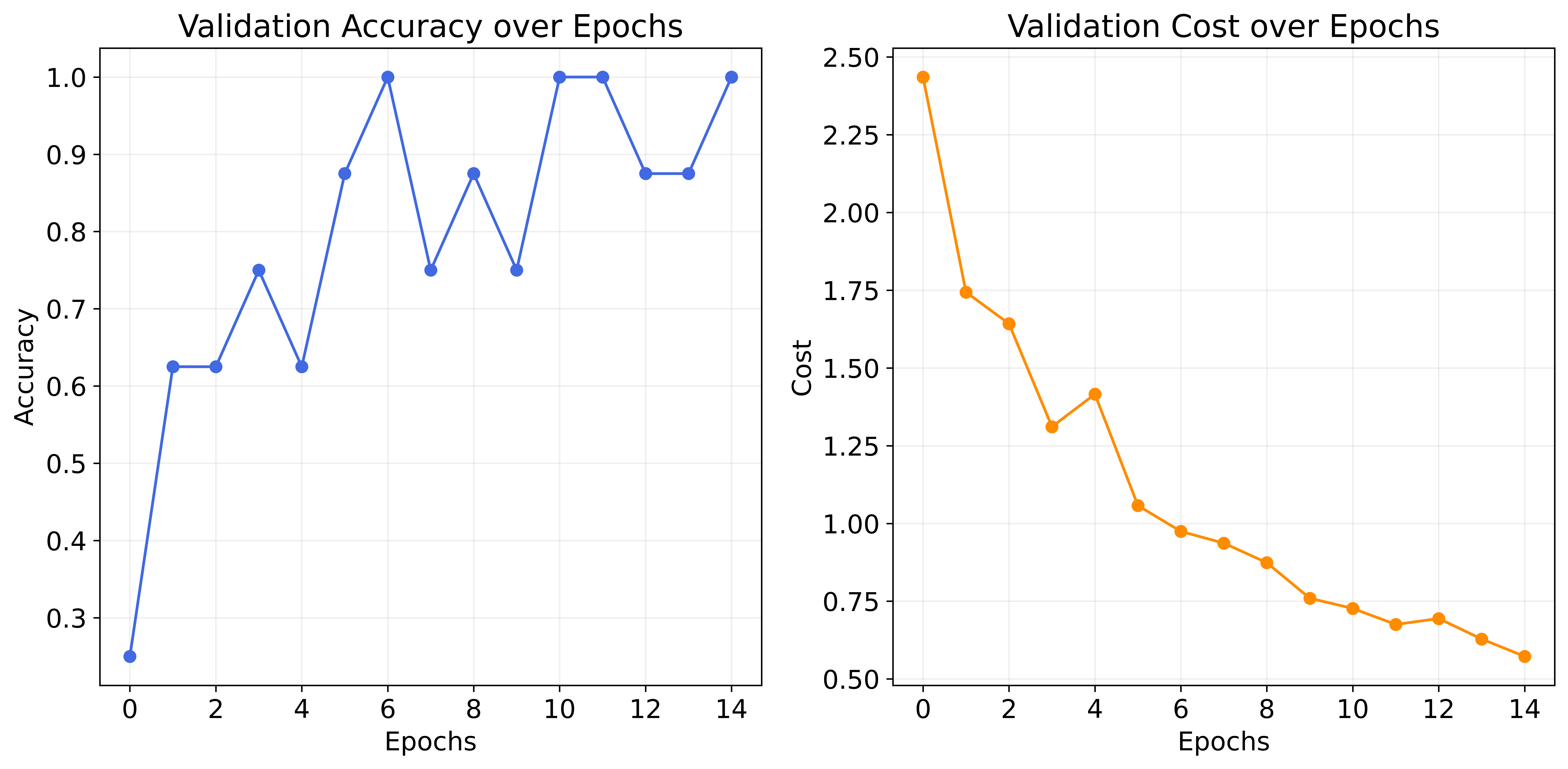}
    \caption{Validation accuracy and cost curves of the Deep VQC model.}
    \label{fig:val_acc_1}
\end{figure}

\begin{figure}
    \centering
    \includegraphics[width=0.8\columnwidth]{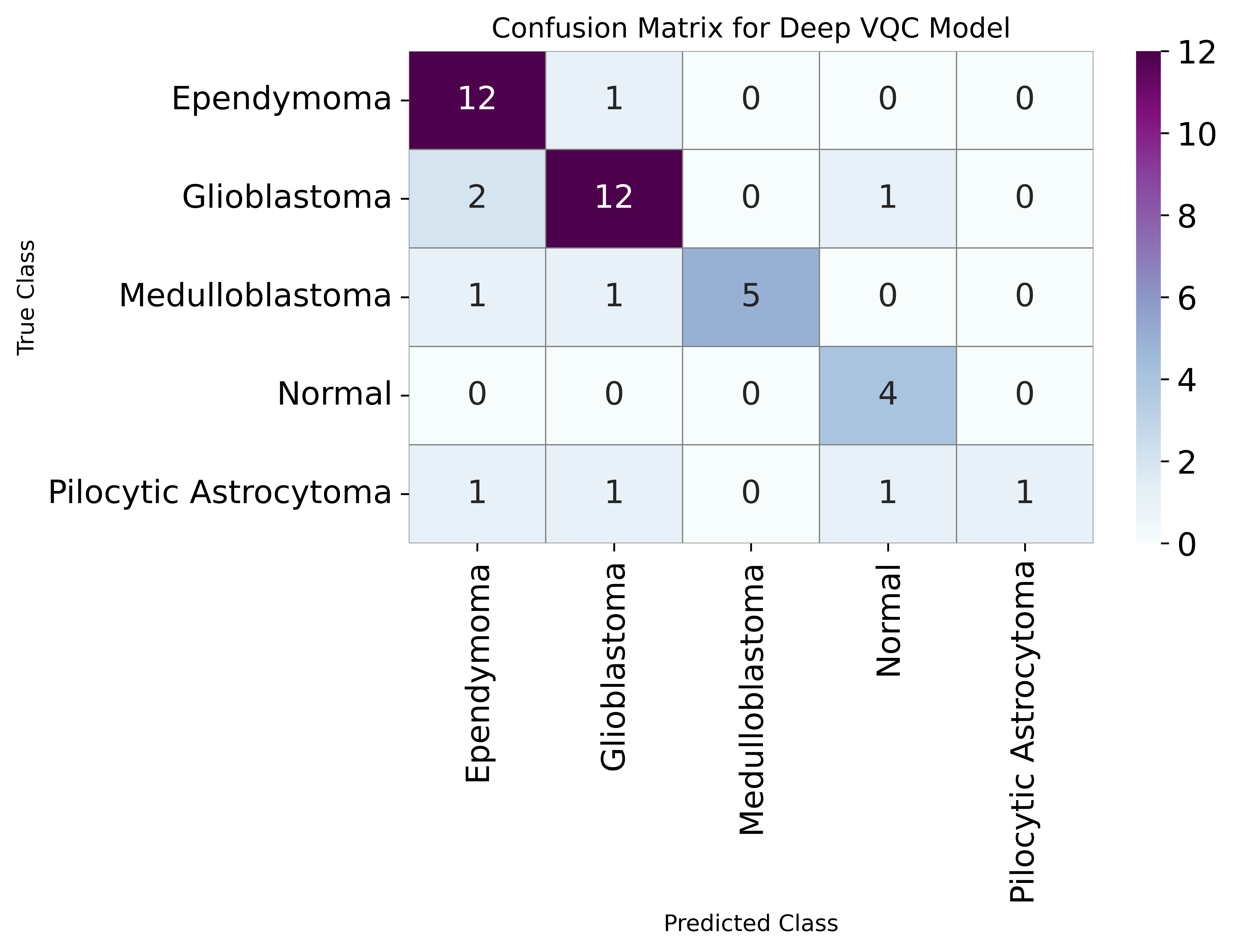}
    \caption{Confusion matrix of the Deep VQC model.}
    \label{fig:conf_mat_1}
\end{figure}

When the Deep VQC model was employed to distinguish four different brain tumor types and normal samples based on microarray data, the obtained training and validation accuracy values were 0.88 and 0.79, respectively, while the corresponding training and validation cost values were 0.63 and 0.76. The training accuracy and training cost curves corresponding to these results are presented in Fig. \ref{fig:train_acc_1}, whereas the validation accuracy and validation cost curves are shown in Fig. \ref{fig:val_acc_1}. For this model, the precision, recall, and F1-score values for each class range from 0.67 to 1, 0.25 to 1, and 0.40 to 0.83, respectively. The confusion matrix of the Deep VQC model is illustrated in Fig. \ref{fig:conf_mat_1}.

\subsection{Performance of the Deep VQC Model with PCA }
\begin{figure}
    \centering
    \includegraphics[width=0.8\columnwidth]{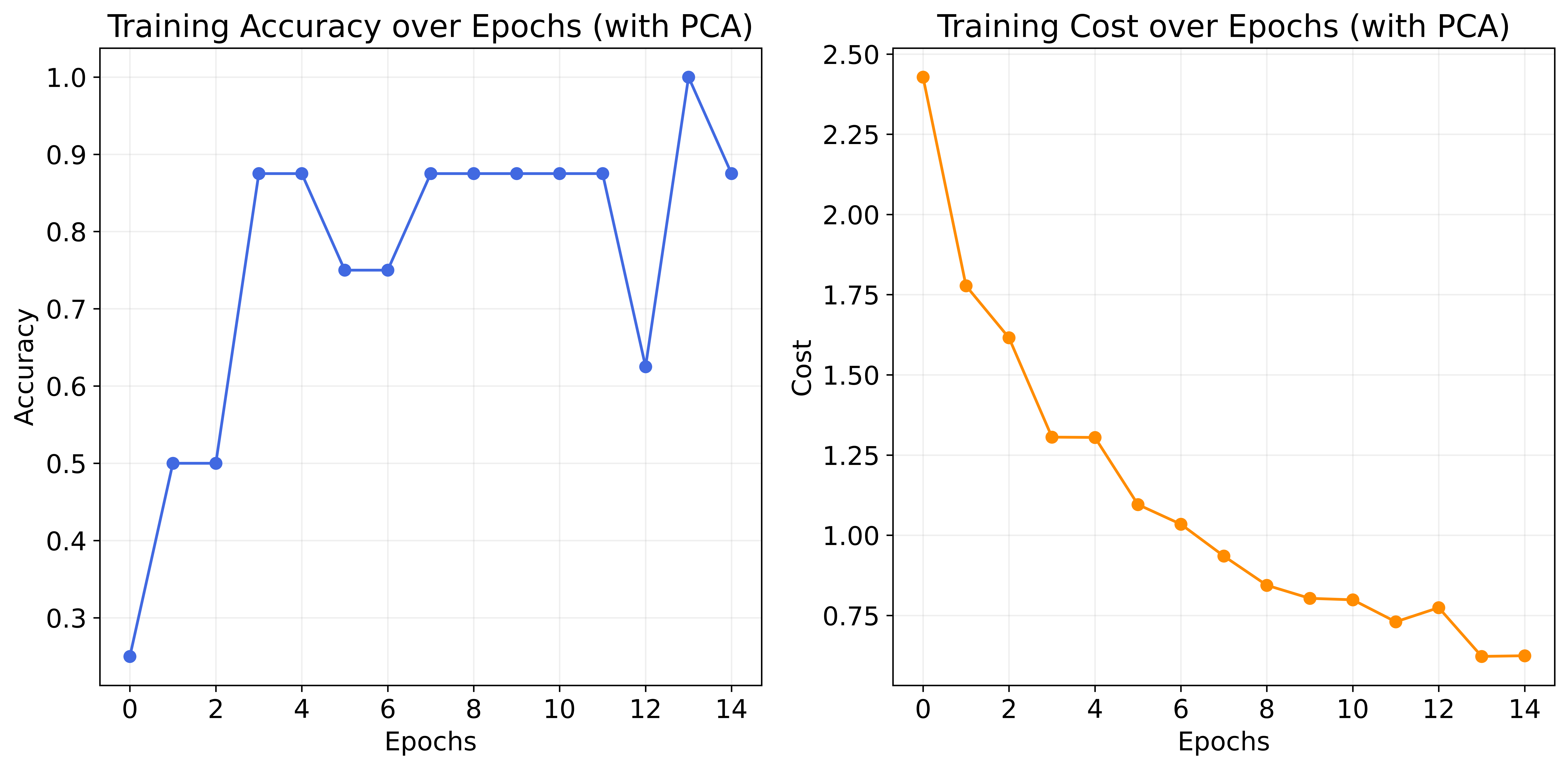}
    \caption{Training accuracy and cost curves of the Deep VQC Model with PCA.}
    \label{fig:train_acc_2}
\end{figure}

\begin{figure}
    \centering
    \includegraphics[width=0.8\columnwidth]{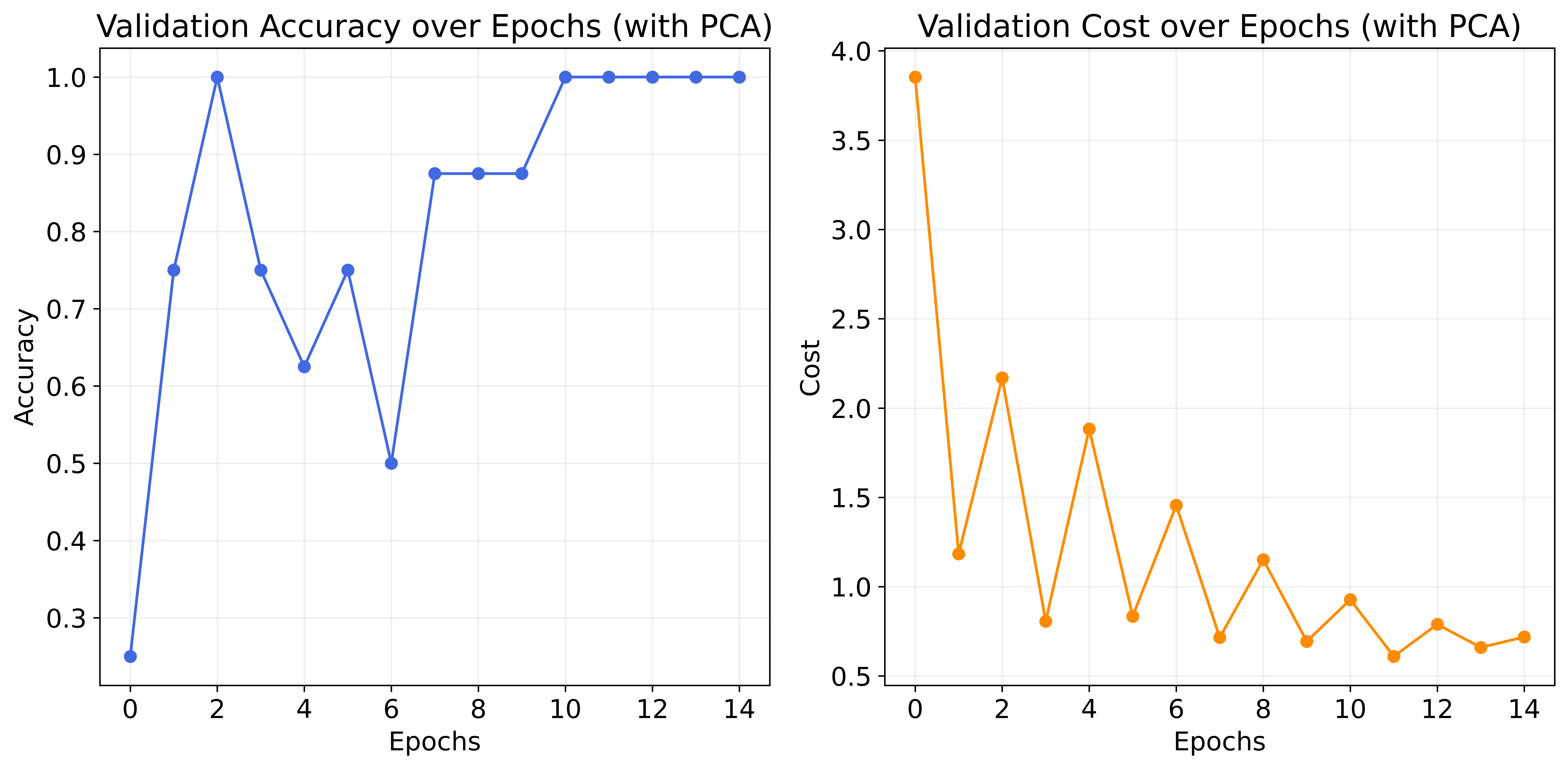}
    \caption{Validation accuracy and cost curves of the Deep VQC Model with PCA.}
    \label{fig:val_acc_2}
\end{figure}

\begin{figure}
    \centering
    \includegraphics[width=0.8\columnwidth]{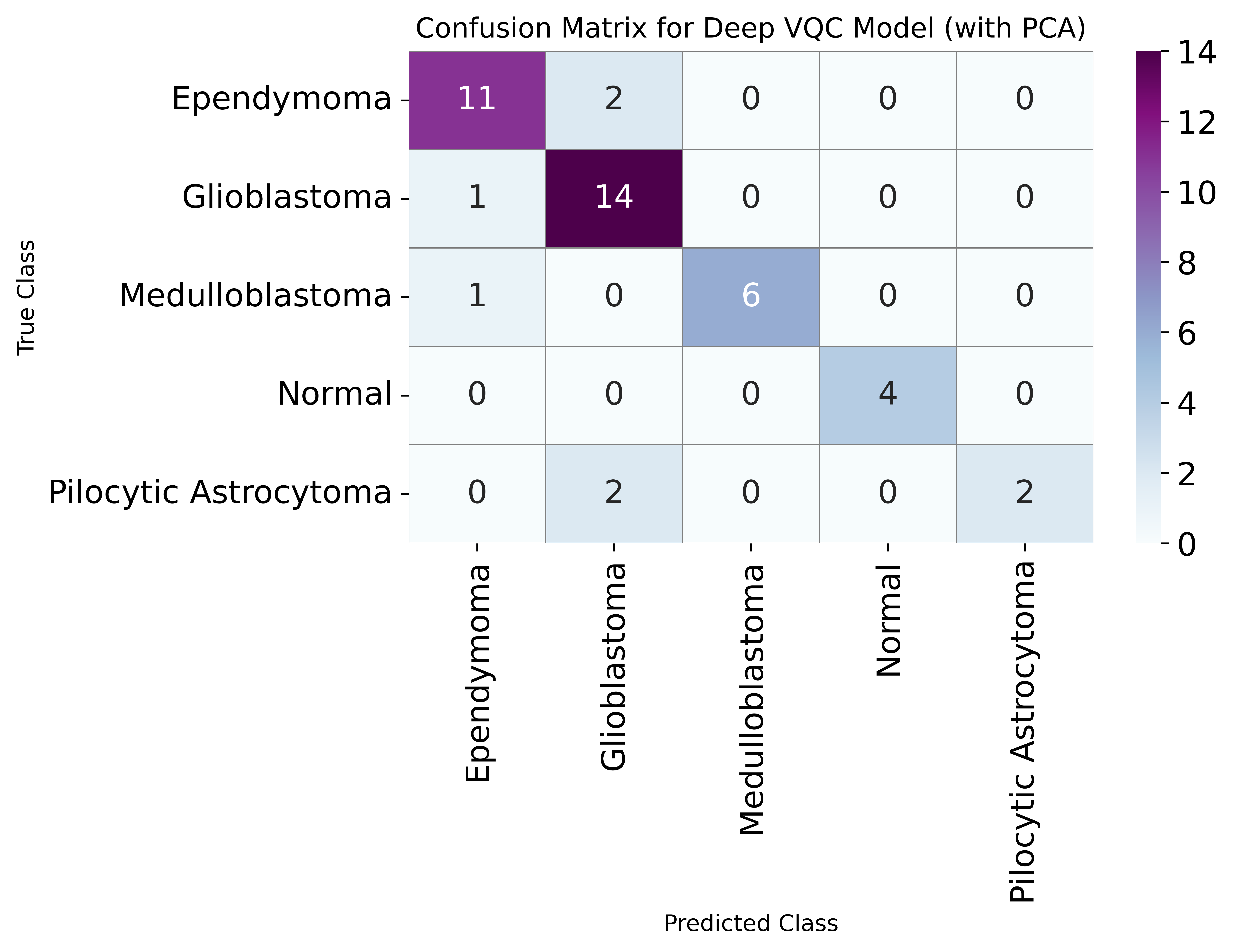}
    \caption{Confusion matrix of the Deep VQC model with PCA.}
    \label{fig:conf_mat_2}
\end{figure}

When PCA was applied during the preprocessing stage of the Deep VQC model, the training and validation accuracy values obtained for distinguishing four different brain tumor types and normal samples were 0.88 and 0.86, respectively, while the corresponding training and validation cost values were 0.62 and 0.76. The training accuracy and training cost curves corresponding to these results are presented on Fig. \ref{fig:train_acc_2}, whereas the validation accuracy and validation cost curves are shown in Fig. \ref{fig:val_acc_2}. For this model, the precision, recall, and F1-score values for each class range from 0.78 to 1, 0.50 to 1, and 0.67 to 0.83, respectively. The confusion matrix of the Deep VQC Model with PCA is illustrated in Fig. \ref{fig:conf_mat_2}.

\subsection{Comparison with Classical Machine Learning Models}
\begin{figure}
    \centering
    \includegraphics[width=0.7\columnwidth]{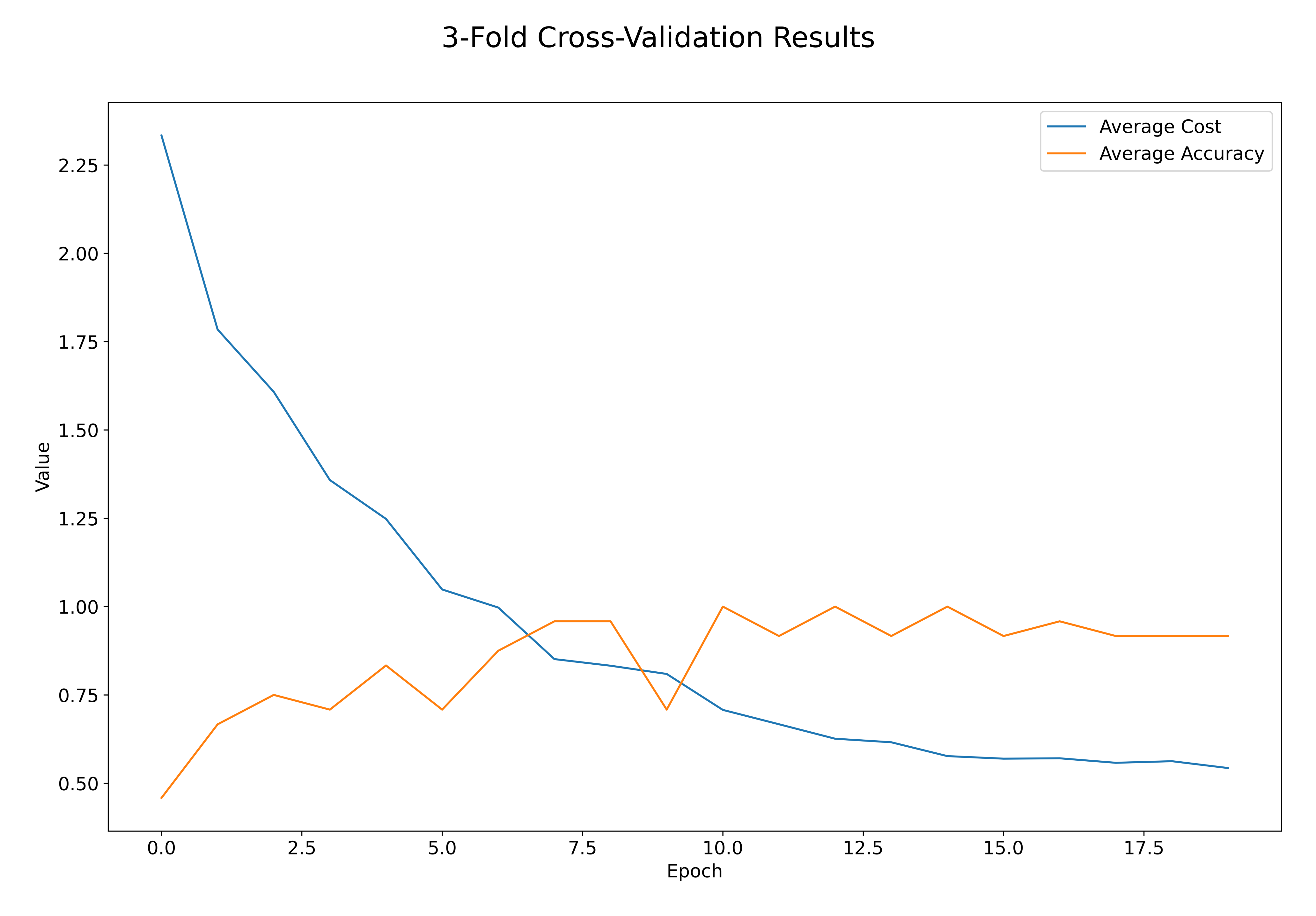}
    \caption{Graphical representation of the average training accuracy and cost values obtained through 3-fold cross-validation of the proposed Deep VQC model}
    \label{fig:cros_val}
\end{figure}

The classification performance of the proposed Deep VQC model was evaluated using the 3-fold cross-validation method. The results demonstrated consistent accuracy values across different data partitions, indicating the robustness and generalization capability of the model. Accordingly, the average classification accuracy obtained from 3-fold cross-validation for the Deep VQC model was 0.85, while the average cost was 0.94. The corresponding training accuracy and cost graphs are presented in Fig.     \ref{fig:cros_val}.

Furthermore, the classification performance of the proposed Deep VQC model in this study was compared with various ML algorithms from the CuMiDa database. The accuracy values for these algorithms are presented in Table \ref{tab1}.

\begin{table}
\caption{Accuracy scores of the proposed Deep VQC, and various ML algorithms reported in the CuMiDa database}
\begin{center}
\begin{tabular}{|c|c|}
\hline
\textbf{Classifier} & \textbf{Accuracy Score} \\
\hline
Deep VQC (Ours)	& 0.85\\
\hline
Decision Tree (DT)	& 0.85 \\
\hline
Hoeffding Tree (HC) & 0.38 \\
\hline
Naive Bayes (NB) &	0.85 \\
\hline
Random Forest (RF)	& 0.91 \\
\hline
K-Nearest Neighbors (KNN) &	0.87 \\
\hline
Multi-Layer Perceptron (MLP) &	0.82 \\
\hline
Support Vector Machine (SVM) &	0.95 \\
\hline
ZeroR	& 0.35 \\
\hline
k-Means	& 0.46 \\
\hline
\end{tabular}
\label{tab1}
\end{center}
\end{table}

\section{Conclusion}
In this study, a deep variational quantum classification model—referred to as Deep VQC—was developed based on the variational quantum classifier (VQC) framework. The model was employed to distinguish between four different brain tumor types (ependymoma, glioblastoma, medulloblastoma, and pilocytic astrocytoma) and samples representing the healthy control group, based on microarray data comprising 54,676 gene expression features. The Deep VQC model includes four main stages: feature mapping, Hardware Efficient Ansatz (HEA), measurement, and optimization. Two distinct HEA circuit designs were implemented and sequentially executed to effectively learn salient features from the input data.
When the full set of 54,676 gene features was directly fed into the Deep VQC model, training and validation accuracies were observed as 0.88 and 0.79, respectively, with corresponding cost values of 0.63 and 0.76. Prior to model training, in the second step, PCA was applied to reduce the dimensionality of the gene features while retaining 95\% of the variance. With this preprocessing, classification accuracies were obtained as 0.88 (training) and 0.86 (validation), while cost values were 0.62 and 0.76, respectively. Comparative results indicate that reducing the input feature space from 54,676 to 65 dimensions led to an approximate 9\% improvement in validation accuracy and a 1.6\% decrease in training cost. These findings suggest that PCA, by preserving essential information, can enhance classification performance and learning efficiency in high-dimensional datasets such as microarray data. However, it should be noted that aggressive dimensionality reduction (approximately 99\%) may lead to limited performance gains and carries the risk of potential information loss. Thus, dimensionality reduction techniques like PCA should be applied with caution and balance, especially in complex and large-scale biological datasets.
In the final stage of the study, the classification performance of the Deep VQC model was compared against the 3-fold cross-validation results of nine traditional ML algorithms reported in the CuMiDa database. Using 3-fold cross-validation, the Deep VQC model achieved an average classification accuracy of 0.85 and an average cost of 0.94. While these values fell behind SVM (0.95), RF (0.91), and KNN (0.87), the model performed comparably to DT (0.85) and NB (0.85), and outperformed MLP (0.82), HC (0.38), ZeroR (0.35), and K-Means (0.46).
In conclusion, despite the current limitations of quantum hardware in the NISQ era, the proposed Deep VQC model demonstrates that competitive classification performance can be achieved on high-dimensional and complex gene expression data. Furthermore, the results highlight the potential of quantum-based models to enhance classification tasks. Future work will aim not only to improve classification accuracy but also to develop quantum AI approaches capable of identifying key gene features critical for distinguishing brain tumor types from microarray data.

\section*{Acknowledgement}
This research used resources of the National Energy Research Scientific Computing Center, a DOE Office of Science User Facility supported by the Office of Science of the U.S. Department of Energy under Contract No. DE-AC02-05CH11231 using NERSC award NERSC DDR-ERCAP0033396.

\bibliographystyle{IEEEtran}
\bibliography{main}

\end{document}